\documentclass[runningheads]{llncs}
\usepackage{graphicx}
\usepackage{comment}
\usepackage{amsmath,amssymb} 
\usepackage{color}
\usepackage{dsfont}
\usepackage{siunitx}
\usepackage{multirow}
\usepackage{booktabs} 
\usepackage{standalone}
\usepackage{xspace}
\usepackage{comment}
\usepackage{multirow}
\usepackage[numbers]{natbib}
\usepackage{multicol}
\usepackage[bookmarks=true]{hyperref}
\usepackage[capitalise]{cleveref} 
\usepackage{etoolbox}
\usepackage{subfig}

\renewcommand{\vec}[1]{\mathbf{#1}}

\newcommand{\myparagraph}[1]{\vspace{3pt}\noindent{\bf #1}}
\newcommand{\etal}{\textit{et al.}}

\newcommand{\I}{\mathcal{I}}
\newcommand{\St}{\mathcal{S}}

\newcommand{\Obj}{\mathcal{O}}
\newcommand{\F}{\mathbf{F}}
\newcommand{\A}{\mathbf{A}}
\newcommand{\V}{\mathbf{V}}

\newcommand{\Y}{\mathcal{Y}}

\begin{document}
\pagestyle{headings}
\mainmatter

\title{Attentional Bottleneck: Towards an Interpretable Deep Driving Network}
\titlerunning{Attentional Bottleneck: Towards an Interpretable Deep Driving Network}
\author{Jinkyu Kim \and Mayank Bansal}
\authorrunning{J. Kim and M. Bansal}
\institute{Waymo Research\\
\email{\{jinkyukim, mayban\}@waymo.com}}

\maketitle

\begin{abstract}
Deep neural networks are a key component of behavior prediction and motion generation for self-driving cars. One of their main drawbacks is a lack of transparency: they should provide easy to interpret rationales for what triggers certain behaviors. We propose an architecture called {\textit{Attentional Bottleneck}} with the goal of improving transparency. Our key idea is to combine visual attention, which identifies what aspects of the input the model is using, with an information bottleneck that enables the model to only use aspects of the input which are important. This not only provides sparse and interpretable attention maps (e.g. focusing only on specific vehicles in the scene), but it adds this transparency at no cost to model accuracy. In fact, we find slight improvements in accuracy when applying Attentional Bottleneck to the ChauffeurNet model, whereas we find that the accuracy deteriorates with a traditional visual attention model.

\end{abstract}

\section{Introduction}
Deep neural networks are powerful function estimators and have been a key component in self-driving software systems \cite{bansal2018chauffeurnet, zeng2019end}. Such networks are, however, notoriously cryptic -- their hidden layer activations may have no obvious relation to the function being estimated by the network. Interpretable models that make deep models more transparent are important for a number of reasons: (1) {\textit{user-acceptance}}: neural network autonomous control is a radical technology that requires a very high-level of user-trust, (2) {\textit{extrapolation}}: users should be able to anticipate what the vehicle will do in most scenarios by understanding the causal behavior of the model, and (3) {\textit{human-vehicle communication}}: communication can be grounded in the vehicle's internal state.

One way of making models transparent is via visual attention~\cite{xu2015show, park2016attentive, kim2017interpretable}.
Visual attention finds spatially varying scalar attention weights $\alpha(x,y) \in [0, 1]$ typically by learning a multi-layer perceptron from a set of input features $\F = \{{\bf{f}}(x,y)\}$. Attended features $\A = \{{\bf{a}}(x,y)\}$ obtained as ${\bf{a}}(x,y) = \alpha(x,y){\bf{f}}(x,y)$ are then used by the model instead of the original features $\F$. The model is trained end-to-end leading the attention weights to link the network's output to its input -- visualizing the weights as a 2D heatmap thus provides insight into the areas of the input image that the network attends to. Furthermore, to be easily interpretable, attention needs to be sparse (i.e. low entropy), while ideally also enhancing the performance of the original model. Unfortunately, given the complexity of the driving task, we find that a straightforward integration of attention maps tends to find all potentially salient image areas, resulting in limited interpretability (e.g. Fig.~\ref{fig:diagram}).

In this work, we manage to achieve sparse and salient attention maps and good final model performance, by attaching attention to a bottlenecked latent representation of the input. However, given the information loss in the bottleneck, we need to provide the model direct access to a subset of dense inputs (e.g. road lane geometry and connectivity information) that are harder to compress. This frees up the bottleneck branch to focus on selecting the most relevant parts of the dynamic input (e.g. nearby objects), while retaining the model performance.

\begin{figure}[!t]
    \centering
        \includegraphics[width=.95\linewidth]{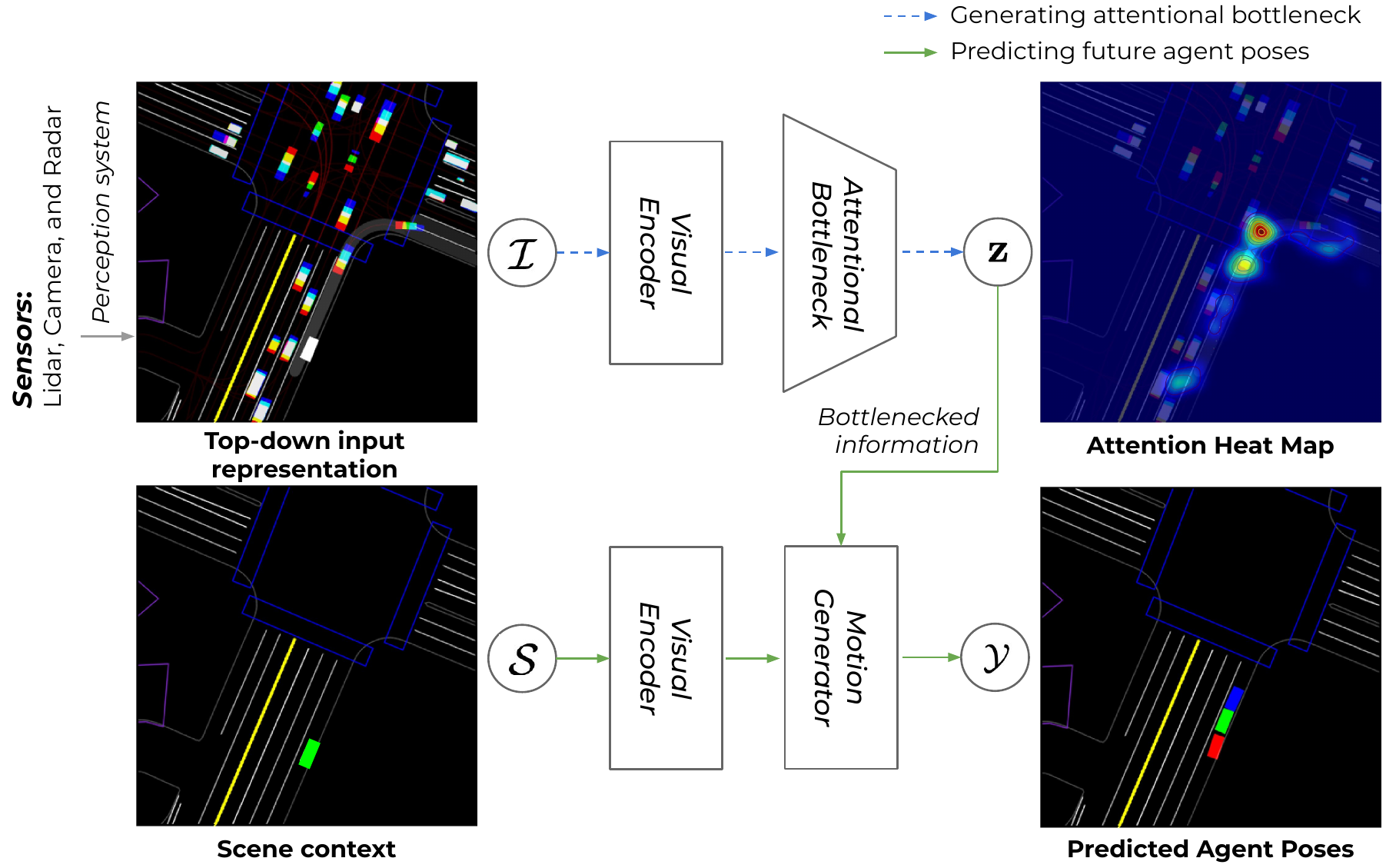}
    \caption{An overview of our interpretable driving model. Our model takes a top-down input representation $\I$ and outputs the future agent poses $\mathcal{Y}$ along with an attention map. An {\em Attentional Bottleneck} encodes the inputs $\I$ to a latent vector $\vec{z}$ while also producing an interpretable attention heat map. The motion generator operates in a partially observable environment using only the dense scene context $\St \subset \I$ along with $\vec{z}$ to predict poses $\mathcal{Y}$.}
    \label{fig:teaser}
\end{figure}

End-to-end driving models that directly process a camera image as input have several scene elements confounded into nearby pixels thus making a separation into dense and sparse input subsets infeasible. Therefore, we focus on improving the interpretability of a driving model that uses a mid-level input representation. This means that instead of directly using low-level sensor data, the model uses higher-level semantic information like objects detected by a perception system. As a proxy for such a network, we work with the recently published {\em ChauffeurNet}~\cite{bansal2018chauffeurnet} model, although the ideas presented are more generally applicable. The inputs $\I$ to this network consist of information about the roadmap, traffic lights, dynamic objects, etc. rendered in separate channels in common top-down view coordinate system around the agent. The model predicts future agent poses $\Y$ in the same top-down view (see Fig.~\ref{fig:inputs}). 

To generate sparser and more interpretable attention maps, we propose an architecture called \textit{Attentional Bottleneck} (Fig.~\ref{fig:teaser}) that combines visual attention with the information bottleneck approach~\cite{tishby2000information} of training deep models through supervised learning~\cite{alemi2016deep, chalk2016relevant, goyal2019infobot}. We define $\vec{z}$ as a bottleneck latent representation of an attention weighted feature encoding $\A_\I = {\boldsymbol\alpha}_\I \cdot \F_\I$ of the input features $\F_\I$. We leverage the mid-level input representation to separate the subset of dense inputs into a set $\St \subset \I$. Conditioned on $\vec{z}$ and $\St$, the motion generator finally predicts the target $\mathcal{Y}$. Our goal is to learn both the attention weighting function ${\boldsymbol\alpha}_\I$ and an encoding $\vec{z}$ that is maximally informative about the target $\mathcal{Y}$. To prevent $\vec{z}$ from being the identity encoding of the inputs and to focus the network on specific areas of causality, we impose an information bottleneck constraint on the complexity of $\vec{z}$ by a pooling operation. We preserve spatial information in the attention map by incorporating a positional encoding step, and encode non-local information by using Atrous convolutions. 

We evaluate our approach on the large-scale ($\approx$ 60 days of continuous driving) dataset from \cite{bansal2018chauffeurnet} and show quantitative and qualitative results illustrating that our generated attention maps result in much sparser (and thus more interpretable) visualization of the internal states than a baseline visual attention model. We also show that our approach improves the motion generation accuracy in contrast to a traditional visual attention model that results in decreased accuracy.

\section{Related Work}
\subsection{Deep Driving Models}
Recently there is growing interest in end-to-end driving models that process raw sensor data to directly output driving controls. Most of these approaches learn a driving policy through supervised regression over observation-action pairs from human drivers. ALVINN (Autonomous Land Vehicle In a Neural Network)~\cite{pomerleau1989alvinn} was the first attempt to train a neural network for the navigational task of road following. Bojarski~\etal~\cite{bojarski2016end} trained a 5-layer ConvNet to predict steering controls only from a dashcam image, while Xu~\etal~\cite{xu2016end} utilized a dilated ConvNet combined with an LSTM so as to predict vehicle's discretized future motions. Hecker~\etal~\cite{hecker2018end} explored the extended model that takes a surround-view multi-camera system, a route planner, and a CAN bus reader. Codevilla~\etal~\cite{codevilla2018end} explored a conditional end-to-end driving model that takes high-level command input (i.e. left, straight, right) at test time. These models show good performance in simple driving scenarios (i.e. lane following). Their behavior, however, is opaque and learning to drive in urban areas remains challenging.

To reduce the complexity and for better interpretability, there is growing interest in end-to-mid and mid-to-mid driving models that produce a mid-level output representation in the form of a drivable trajectory by consuming either raw sensor or an intermediate scene representation as input. Zeng~\etal~\cite{zeng2019end} introduced an end-to-mid neural motion generator, which takes Lidar data and an HD map as inputs and outputs a future trajectory. This model also detects 3D bounding boxes as an intermediate representation. Bansal~\etal~\cite{bansal2018chauffeurnet} introduced {\em ChauffeurNet}, a mid-to-mid model that takes advantage of separate perception and control components. Using a top-down representation of the environment and intended route as input, the model outputs a driving trajectory that is consumed by a controller, which then translates it to steering and acceleration. Recent works~\cite{chai2019multipath, wang2019monocular} also suggest that such a top-down scene representation can successfully be used to learn high-level semantic information. In this work, we focus on improving the explainability of such a model.

\subsection{Visual Explanations}
Explainability of deep neural networks has seen growing interest in computer vision and machine learning~\cite{gunning2017explainable}. In landmark work, Zeiler~\etal~\cite{zeiler2014visualizing} utilized deconvolution layers to visualize the internal representation of a ConvNet. Bojarski~\etal~\cite{bojarski2016visualbackprop} developed a richer notion of contribution of a pixel to the output, while other approaches~\cite{zhou2016learning, selvaraju2017grad} have explored synthesizing an image causing high neuron activations. However, a difficulty with de-convolution based approaches is the lack of a formal notion of contribution of spatially-extended features (rather than pixels). 

Attention-based approaches~\cite{xu2015show, park2016attentive} have been increasingly employed for improving a model's ability to explain by providing spatial attention maps that highlight areas of the image that the network attends to. Kim~\etal~\cite{kim2017interpretable} utilize an attention model followed by additional salience filtering to show regions that causally affect the output. To reduce the complexity of explanations, Wang~\etal~\cite{wang2018deep} introduce an instance-level attention model that finds objects (i.e. cars and pedestrians) that the network needs to pay attention to. Such attention may be more intuitive and interpretable for users to understand the model's behavior. However, the model needs to take the whole input context as an additional input, which may compromise the causality of the attention -- explanations may not represent causal relationships between the system's input and its behavior. To preserve the causality, we use a top-down representation of the environment as an input, which consists of information around the agent rendered in separable channels. 

Another notable approach is the work by Chen~\etal~\cite{chen2015deepdriving}, which defined human-interpretable intermediate features such as the curvature, deviation to neighboring lanes, and distances from the front-located vehicles. A CNN is trained to produce these features from an image, and a simple controller maps them to steering angle. Similarly, Sauer~\etal~\cite{sauer2018conditional} proposed conditional affordance learning approach that maps visual inputs to intermediate representations conditioned on high-level command input. However, the intermediate feature descriptors provide a limited and ad-hoc vocabulary for explanations. Zeng~\etal~\cite{zeng2019end} co-trained a perception model that provides bounding boxes of dynamic objects, which are then used as an intermediate and interpretable feature. Here, we instead take full advantage of existing well-established perception systems with different sensor sources (i.e. Lidar, Radar, and Camera) via a mid-level input representation. This reduces the complexity of the driving network and employs more reliable perception outputs.

There is also a growing effort on textual explanations that justify the decisions that were made and explain the ``why'' in a natural language~\cite{hendricks2018grounding, park2016attentive, kim2018textual}. However, textual explanations are often rationalizations -- explanations that justify the system's behavior in a post-hoc manner -- and are less helpful with understanding the causal behavior of the model. In this work, we focus on improving attention-based mechanism -- to provide introspective explanations that are based on the system's internal state and represent causal relationships between the system's input and its behavior. 

\begin{figure}[!t]
    \centering   
    \subfloat{
        \includegraphics[width=.55\linewidth]{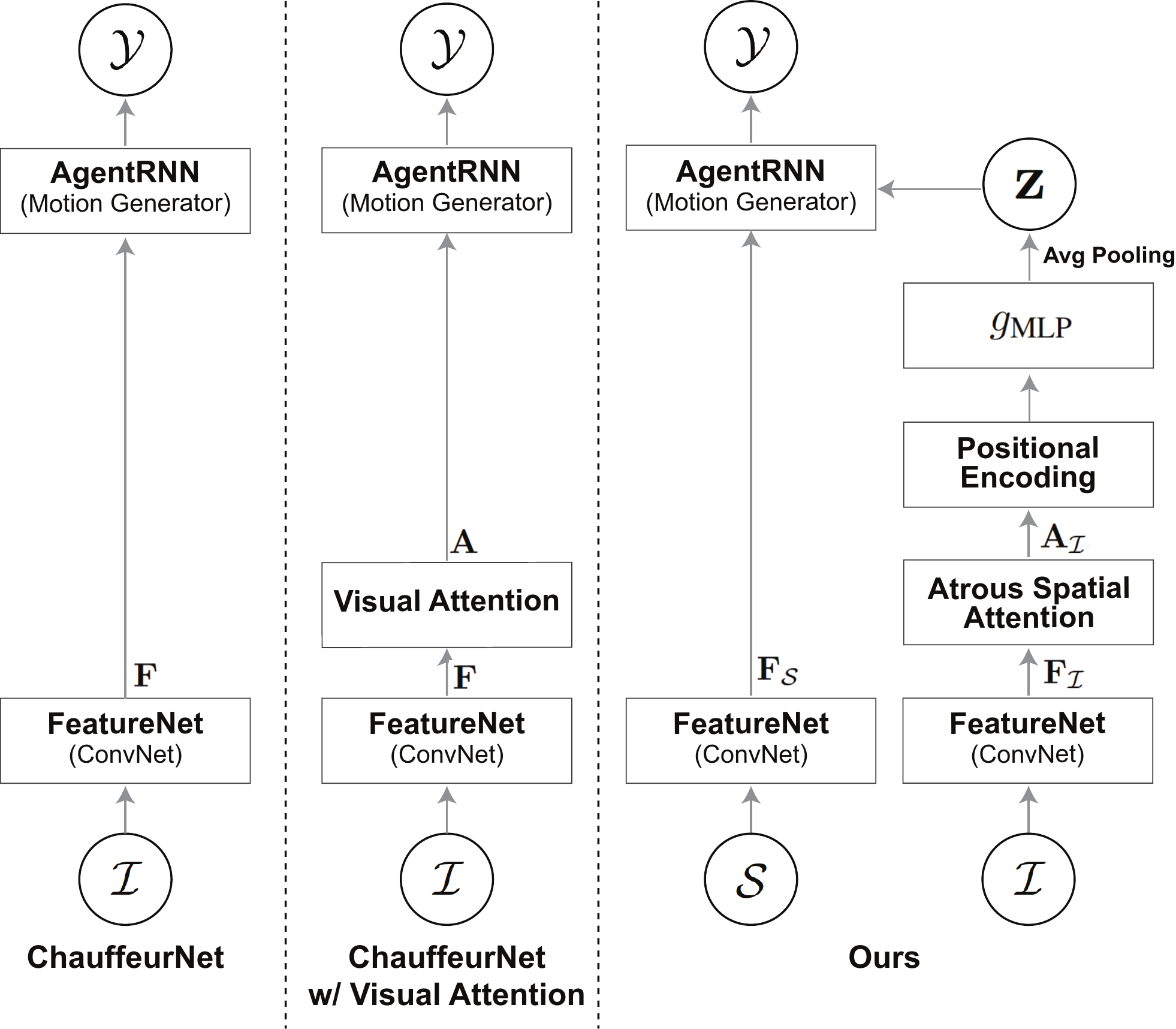}}
    \subfloat{
        \includegraphics[width=.34\linewidth]{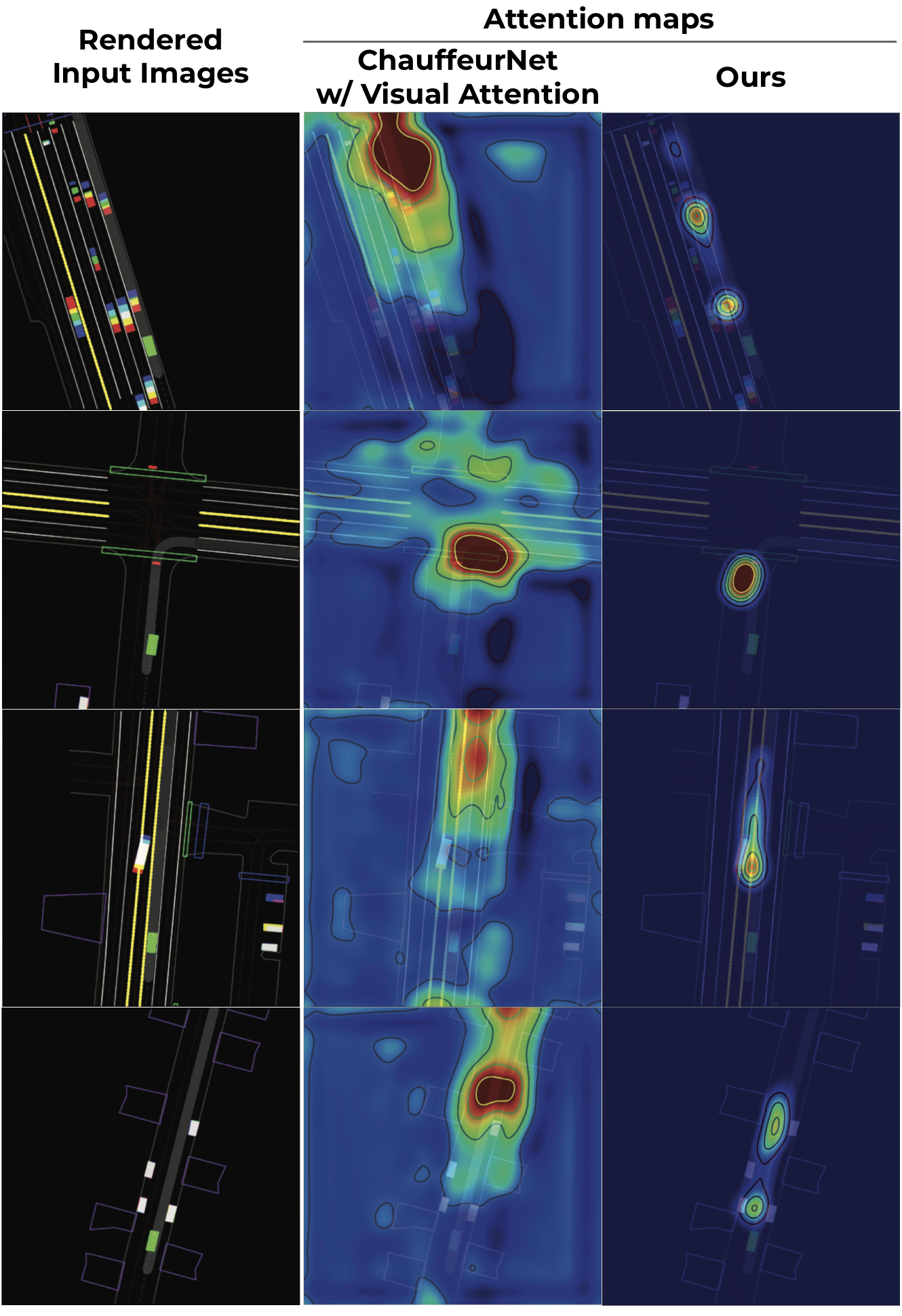}}
    \caption{(left) Attentional Bottleneck design compared with a baseline visual attention model applied to ChauffeurNet. (right) Comparison of attention maps from our model against those from a baseline visual attention model. Note that our heatmaps are much sparser and thus more interpretable.}
    \label{fig:diagram}\vspace{-0.5em}
\end{figure}


\section{Mid-to-mid Driving Model with Visual Attention}
\begin{figure}[!t]
    \centering
        \includegraphics[width=\linewidth]{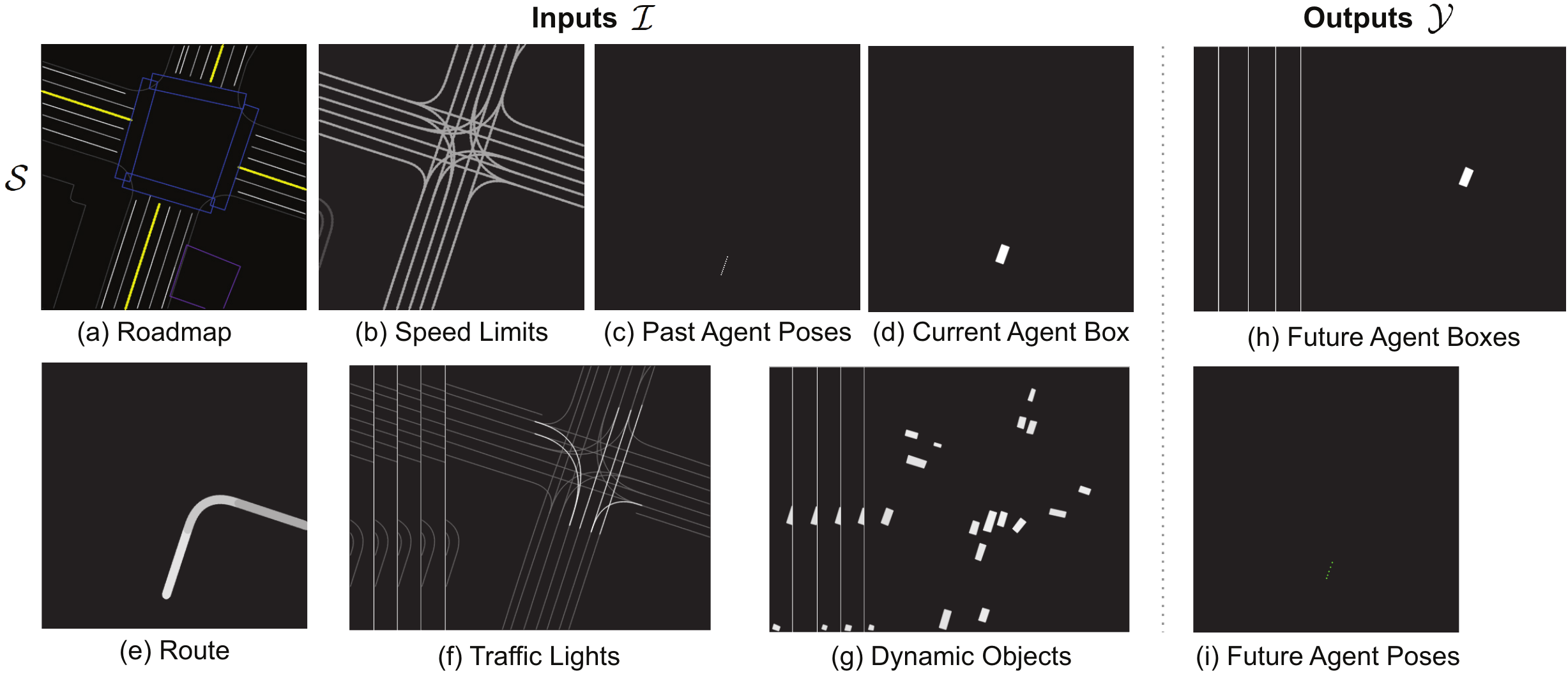}
    \caption{Top-down rendered inputs $\I$ (left) and outputs $\Y$ (right) for the {\em ChauffeurNet} model. The subset of dense scene context inputs $\St$ are shown in the top row.}
    \label{fig:inputs}\vspace{-1.0em}
\end{figure}

\subsection{ChauffeurNet}
Bansal~\etal~\cite{bansal2018chauffeurnet} introduced a mid-to-mid driving network called {\em ChauffeurNet} that recurrently predicts future poses of the agent by processing a top-down representation of the environment as an input. For completeness, we summarize some of the key details of the paper here. The input to the neural network consists of a set of images $\I$ of size $W \times H$ pixels rendered into a top-down view coordinate system that is fixed relative to the current location of the agent. As shown in Fig.~\ref{fig:inputs}, the set $\I$ contains: (a) Roadmap: a 3-channel image with a rendering of color coded lanes, stop signs, cross-walks, etc. (b) Speed limit: a gray-scale image with lane centers color coded in proportion to their known speed limit. (c) Past agent poses: the ego-vehicle's past poses rendered as a trail of points. (d) Current agent box: the current agent represented by a full bounding box. (e) The intended route. (f) Traffic lights: a gray-scale image where each lane center is color coded to reflect different traffic light states (red light: brightest gray level, green light: darkest gray level). (g) Dynamic objects: a gray-scale image that renders all the potential dynamic objects (vehicles, cyclists, pedestrians) as oriented boxes. Both (f) and (g) are a sequence of 5 images reflecting the environment state over the past 5 timesteps.

\myparagraph{Visual Encoder (FeatureNet).}
In the {\em ChauffeurNet} model, the rendered inputs $\I$ are fed to a large-receptive field convolutional
{\em FeatureNet} with skip connections, which outputs features
$\F$ that capture the environmental context and the intent. This feature ${\F}$ (of size $w$$\times$$h$$\times$$d$) contains a set of $d$-dimensional latent vectors over the spatial dimension, i.e. ${\F} = \{ {\bf{f}}_{1}$, ${\bf{f}}_{2}$, $\ldots$, ${\bf{f}}_{l}\}$, where ${{\bf f}_i}\in\mathcal{R}^{d}$ and $l$ (= $w\times h$) is the spatial dimension of the extracted features. Selecting a subset of these feature slices will allow the attention model to selectively attend to different parts of input images.

\myparagraph{Motion Generator (AgentRNN).}
The feature encoding ${\F}$ is fed to a recurrent neural network ({\em AgentRNN}) which predicts the outputs $\Y$ consisting of the next
point $\vec{p}_{k}$ on the driving trajectory, and the agent bounding
box heatmap $B_{k}$, conditioned on the features $\F$, the iteration number $k \in \{1,\ldots,N\}$, 
the memory $M_{k-1}$ of past predictions from {\em AgentRNN}, and the agent bounding box heatmap $B_{k-1}$
predicted in the previous iteration.
\begin{equation}\label{eq:AgentRNNiter}
\vec{p}_{k},B_{k}=\textrm{AgentRNN}(k,\F,M_{k-1},B_{k-1})
\end{equation}

\subsection{ChauffeurNet with Visual Attention}
One way of making models interpretable is via visual attention~\cite{park2016attentive, kim2017interpretable}. These models provide introspective (visual) explanations by filtering out non-salient image regions -- the remaining (attended) regions potentially have a causal effect on the output. The goal of visual attention is to find an attended feature ${\A}=\{{\bf{a}}_{1},{\bf{a}}_{2},\dots,{\bf{a}}_{l}\}$, where ${\bf{a}}_{i}\in\mathcal{R}^{d}$ from the original feature $\F$. They utilize a deterministic soft attention mechanism that is trainable by standard back-propagation methods, which thus has advantages over a hard stochastic attention mechanism that requires reinforcement learning. As discussed by several works~\cite{xu2015show, kim2017interpretable}, the attended features can be computed as ${\bf{a}}_{i} = \pi(\alpha_{i}, {\bf{f}}_{i}) = \alpha_{i}{\bf{f}}_{i}$ for $i=\{1,2,\dots,l\}$, where $\alpha_i$ are scalar attention weights in $[0, 1]$ satisfying $\sum_{i}{\alpha_{i}}=1$. These weights are estimated from the input features $\F$ typically by a multi-layer perceptron, i.e. $\alpha_{i}=f_{\textnormal{MLP}}({\bf{f}}_{i})$ where the parameters of $f_{\textnormal{MLP}}$ are learned as part of training the entire model end-to-end. Since the attention weights vary spatially and depend on the input (via the features $\F$), they can be visualized as an {\em attention heatmap} aligned with the input image, with brighter regions reflecting areas salient for the task. 

To allow us to explain the driving decisions made by ChauffeurNet, we apply this vanilla visual attention approach by replacing the original features $\F$ in Eq.~\ref{eq:AgentRNNiter} with the attended features $\A$ as shown in Fig.~\ref{fig:diagram} (left). As shown in Fig.~\ref{fig:diagram} (right), this approach generates vague and verbose attention maps which do not add to the interpretability of the model. Therefore, we use this approach as a baseline for our ``Attentional Bottleneck'' approach. 

\section{Attentional Bottleneck}\label{ss:bottleneck}
We propose a novel architecture called {\em Attentional Bottleneck} with a focus on generating sparse and fine-grained visual explanations. We encode the environment $\I$ through an information bottleneck that serves to restrict information in the input to only the most relevant parts of the input, and thus allows the driving model to focus on specific features in the environment. We tie this feature selection to the spatial distribution of features by employing a spatial attention mechanism before the bottleneck. While the driving task involves focusing on specific objects and entities in the scene for the immediate driving decisions, humans also employ a holistic understanding of some elements of the environment. For example, humans are aware of the overall map of the environment through visual scanning or through looking at a navigation app. We find that compressing this kind of dense information through the bottleneck either leads to dense attention maps or degrades the model performance. Therefore, we leverage the mid-level separable input representation and provide the model full access to a subset of inputs $\St \subset \I$ containing the dense context about the environment, through a separate branch. This frees up the bottleneck branch to focus on specific parts of the input (e.g. specific objects) making the attention map sparser and more interpretable.

As shown in Fig.~\ref{fig:diagram}, our modified {\em ChauffeurNet} model consists of a dense input encoder branch and an attentional bottleneck branch providing encoded input features to the {\em AgentRNN}.

\myparagraph{Grounding Attentional Bottleneck into AgentRNN.}
Like the baseline model, the inputs $\I$ are first encoded into features $\F_\I$ by the {\em FeatureNet} network. To capture non-local information, we propose an Atrous Spatial Attention layer that computes the attention weights $\boldsymbol\alpha_\I$ and outputs the attended features $\A_\I$. The attended features are depth-concatenated with a positional encoding $\V$ followed by a multi-layer perceptron $g_{\textnormal{MLP}}$, and an average pooling layer to generate the final bottleneck representation $\vec{z}$.
\begin{equation} 
\vec{z}=\sum_{i=1}^l{g_{\textnormal{MLP}}([\vec{a}_i; \vec{v}_i])}
\label{eq:bottleneck}
\end{equation}

The dense scene context inputs $\St$ are similarly encoded into features $\F_\St$ using another {\em FeatureNet} network with identical architecture. We modify {\em AgentRNN} to incorporate the bottleneck vector by concatenating it with each of the features $\vec{f}_i \in \F_\St$:
\begin{align}
\vec{p}_{k},B_{k}=\textrm{AgentRNN}(k,\F_\St,{\bf z},M_{k-1},B_{k-1})\label{eq:AgentRNNattn}
\end{align}

We discuss the Atrous Spatial Attention and the positional encoding stages in the following paragraphs and present ablation results for these blocks in the experiments.

\begin{figure}[!t]
    \centering
        \includegraphics[width=\linewidth]{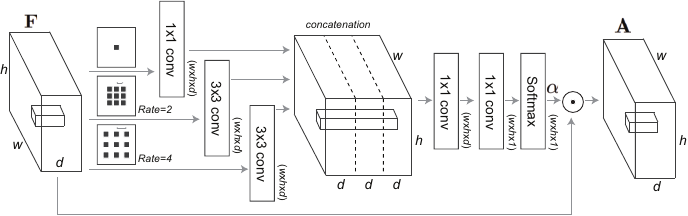}
    \caption{Atrous Spatial Attention Block. We apply three parallel Atrous convolutions with different atrous rates. The resulting features from all three branches are then concatenated and fed into 1x1 convolution and softmax layers to generate the attention weights $\alpha$.}
    \label{fig:atrous_spatial_attention_w_perceptionRNN}\vspace{-1.3em}
\end{figure}

\myparagraph{Atrous Spatial Attention.}
Attention models are typically applied to features generated by the last layer of a convolutional encoder. Attention weights for each spatial location are usually computed independently, allowing them to capture local information around the corresponding specific spatial location (e.g. ``there is a pedestrian running''). However, we argue that the attention model also needs to capture non-local information especially for the driving task (e.g. ``there is a pedestrian running towards the crosswalk ahead''). Seo~\etal~\cite{seo2016progressive} explored using $3\times3$ convolution to consider local context in generating attention maps. Here, we advocate using Atrous convolution (also known as dilated convolution), which has been shown to be effective for accurately capturing semantic information at an arbitrary scale~\cite{chen2017rethinking}.

As shown in Fig.~\ref{fig:atrous_spatial_attention_w_perceptionRNN}, we apply three parallel Atrous convolutions with different rates on top of the feature map ${\F}_{\I}$. For implementation, we closely follow the work by Chen~\etal~\cite{chen2017rethinking}. Specifically, our atrous convolution layers include a 1$\times$1 convolution, and two 3$\times$3 convolutions with rates 2 \& 4 with $d$ filters and batch normalization. The resulting features from all the branches are then concatenated and fed into another 1$\times$1 convolution to generate the attention logits. A spatial softmax yields normalized attention weights $\alpha$.

\begin{figure*}[!t]
    \centering
    \includegraphics[width=.95\linewidth]{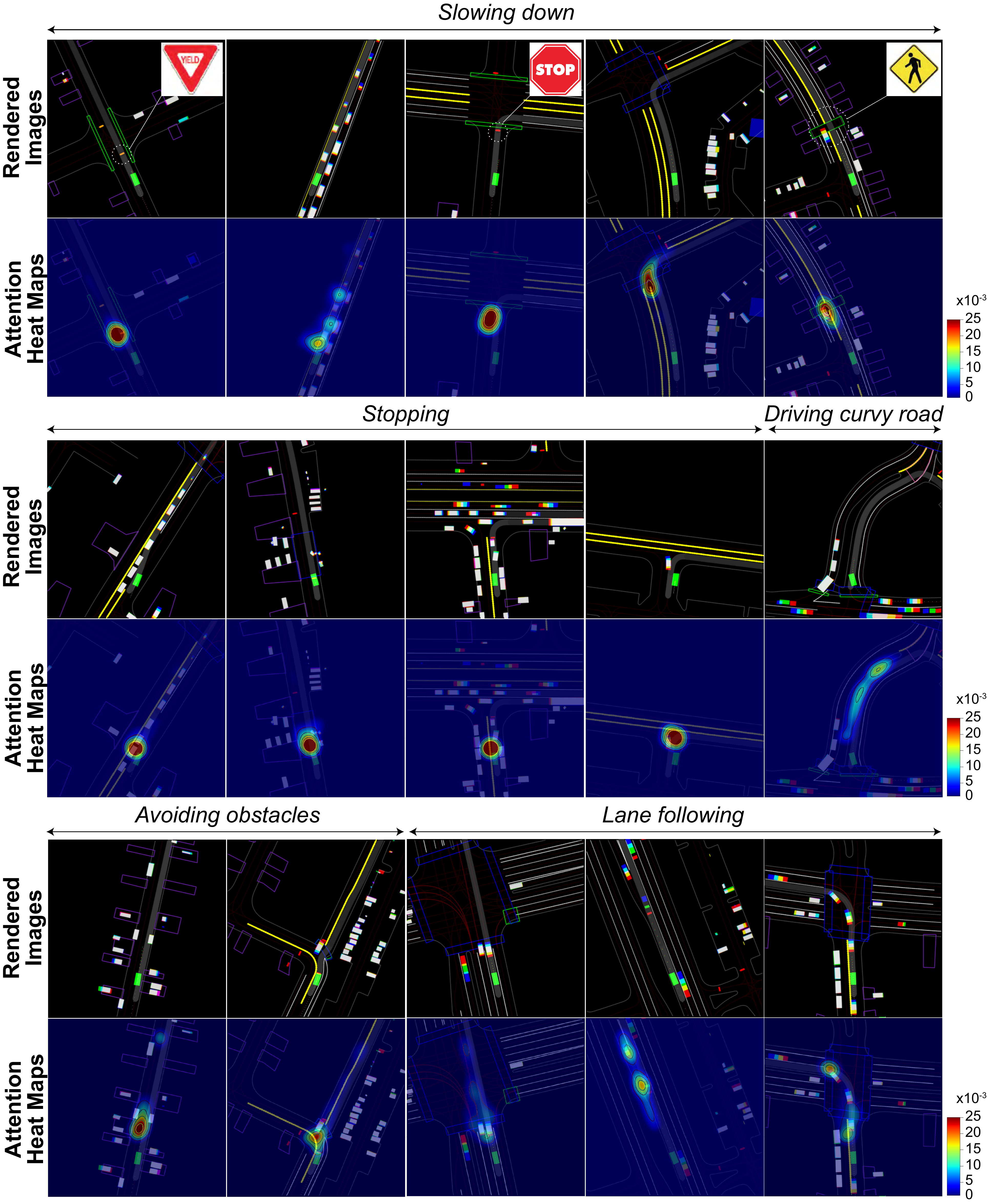}
    \caption{We provide typical examples of attention heat maps in diverse driving scenarios. Our model attends to driving-related visual cues like highlighting stop/yield signs, crosswalks or cars ahead that cause braking, road contours on curved roads, or multiple pinch points from parked cars on narrow roads.}
    \label{fig:attn}
\end{figure*}

\myparagraph{Positional Encoding.} 
As shown in Equation~\ref{eq:bottleneck}, to obtain the latent bottleneck vector $\vec{z}$, we use spatial summation with the attended features ${\A}_{\I}$, which removes the positional information. To preserve this information, we append a spatial basis to the feature $\A_{\I}$. Following Vaswani~\etal~\cite{vaswani2017attention} and Parmar~\etal~\cite{parmar2018image}, we generate a spatial basis ${\V}$ (of the same dimension as ${\F}$) that contains $d$-dimensional vectors ${\V} = \{{\bf{v}}_{1}, {\bf{v}}_{2}, \ldots, {\bf{v}}_{l}\}$, where ${\bf{v}}_{i}\in\mathcal{R}^{d}$. Each vector ${\bf{v}}_{i}$ encodes positional information about the spatial location $(x_i,y_i)$ using four types of Fourier basis functions viz. $\sin(x_i/f_{u})$, $\cos(x_i/f_{u})$, $\sin(y_i/f_{u})$, and $\cos(y_i/f_{u})$, where $f_{u}={1000}^{u}$ is the spatial wavelength with channel index $u=\{0, 4/d, 8/d, \dots, d/d\}$. Each positional encoding feature ${{\bf v}_{i}}$ is then concatenated with the corresponding attended feature ${{\bf a}_i}\in{\A_{\I}}$ as shown in Equation~\ref{eq:bottleneck}.

\section{Experiment}
\begin{figure}[!t]
    \begin{center}
        \includegraphics[width=\linewidth]{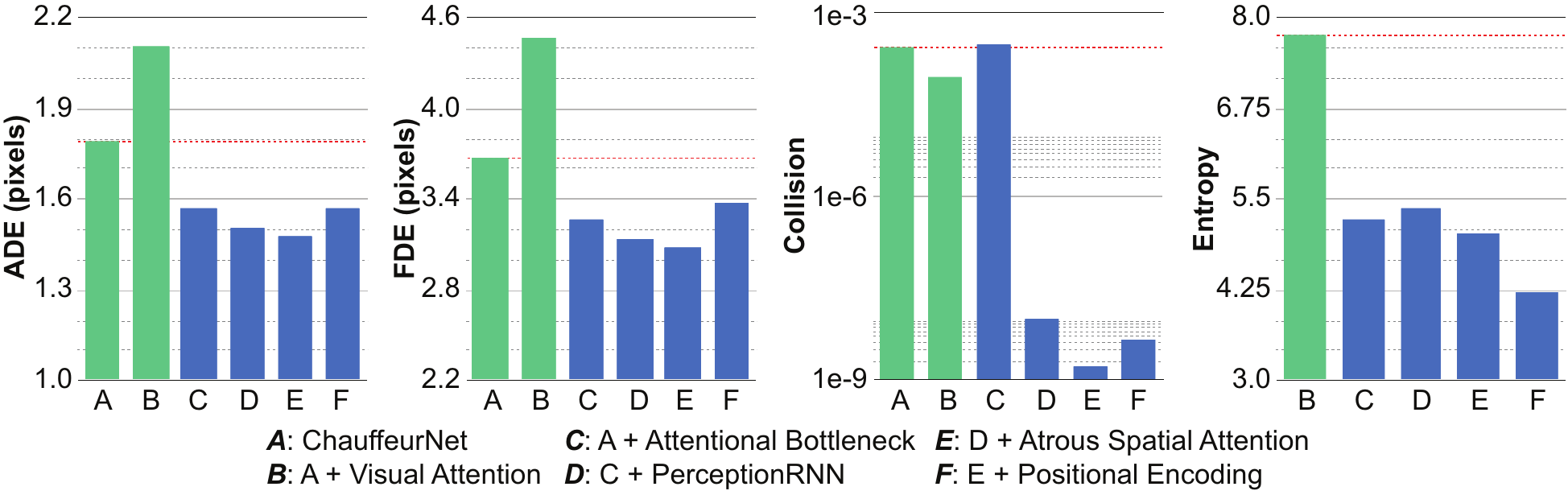}
    \end{center}
    \caption{Comparison of motion generation performance and attention map sparsity between baseline ChauffeurNet, visual attention and Attentional Bottleneck ablation designs.}
    \label{fig:ablation} \vspace{-1.3em}
\end{figure}

\subsection{Dataset}
We use the large-scale dataset from \cite{bansal2018chauffeurnet} that contains over 26 million expert driving examples amounting to about 60 days of continuous driving. Data has been collected by a vehicle instrumented with multiple sensors (i.e. cameras, lidar, radar). A reliable perception system provides accurate environmental descriptions including dynamic objects (i.e. vehicles and pedestrians) and traffic light states. Along with perception, the dataset also provides: (i) the prior map of the environment (i.e. roadmap), (ii) vehicle pose information, and (iii) the speed-limits. The input field of view is $80m\times80m$ (a resolution of $400\times400$ pixels in image coordinates) and the effective forward sensing range of the ego-vehicle is $64$ meters.

\subsection{Training and Evaluation Details}
We trained our models end-to-end with Adam optimization~\cite{kingma2014adam} using exponential decaying learning rates and random initialization (i.e. no pre-trained weights), with ChauffeurNet's default losses. Our FeatureNet creates features ${\F}$ with dimensions $50\times50\times128$ which lead to $50\times50$ attention maps that are upsampled to the input resolution of $400\times400$ by a pyramid expansion step. Note that we use ChauffeurNet's default losses and training strategy~\cite{bansal2018chauffeurnet} to train our model end-to-end. The losses consist of pure imitation losses (i.e. agent position, heading, box, meta prediction loss)  as well as environment losses to provide better generalization (i.e. collision loss, on-road loss, geometry loss, and auxiliary losses). For quantitative evaluation, we use the following metrics:

\myparagraph{ADE and FDE.} To quantitatively evaluate motion generation performance, we use two widely-used (Euclidean distance-based) metrics: (i) the average displacement error (ADE) $\frac{1}{K}\sum_{k=0}^{K} ||\hat{{\bf{p}}}_k-{\bf{p}}_k^{gt}||_{2}$, and (ii) the final displacement error (FDE) $||\hat{{\bf{p}}}_K-{\bf{p}}_K^{gt}||_{2}$, where $K=10$ is the total number of predicted waypoints, and the superscript $gt$ denotes the ground-truth values.

\myparagraph{Collision.} We also use the collision rate by measuring the potential overlap of the predicted agent box with the ground-truth boxes of all other objects in the scene, i.e.  $\frac{1}{K}\sum_{k=0}^{K}\sum_{i,j}B_{k-1}^{gt}(i,j)B_{k}(i,j).$

\myparagraph{Entropy $S(\alpha)$.} To measure the sparseness of the generated attention maps, we measure the entropy of the generated attention heat map $\alpha$, i.e. $S(\alpha)=-\sum_{i=1}^{l} \alpha_{i}\log \alpha_{i}$.

\begin{figure}[!t]
    \begin{center}
        \includegraphics[width=\linewidth]{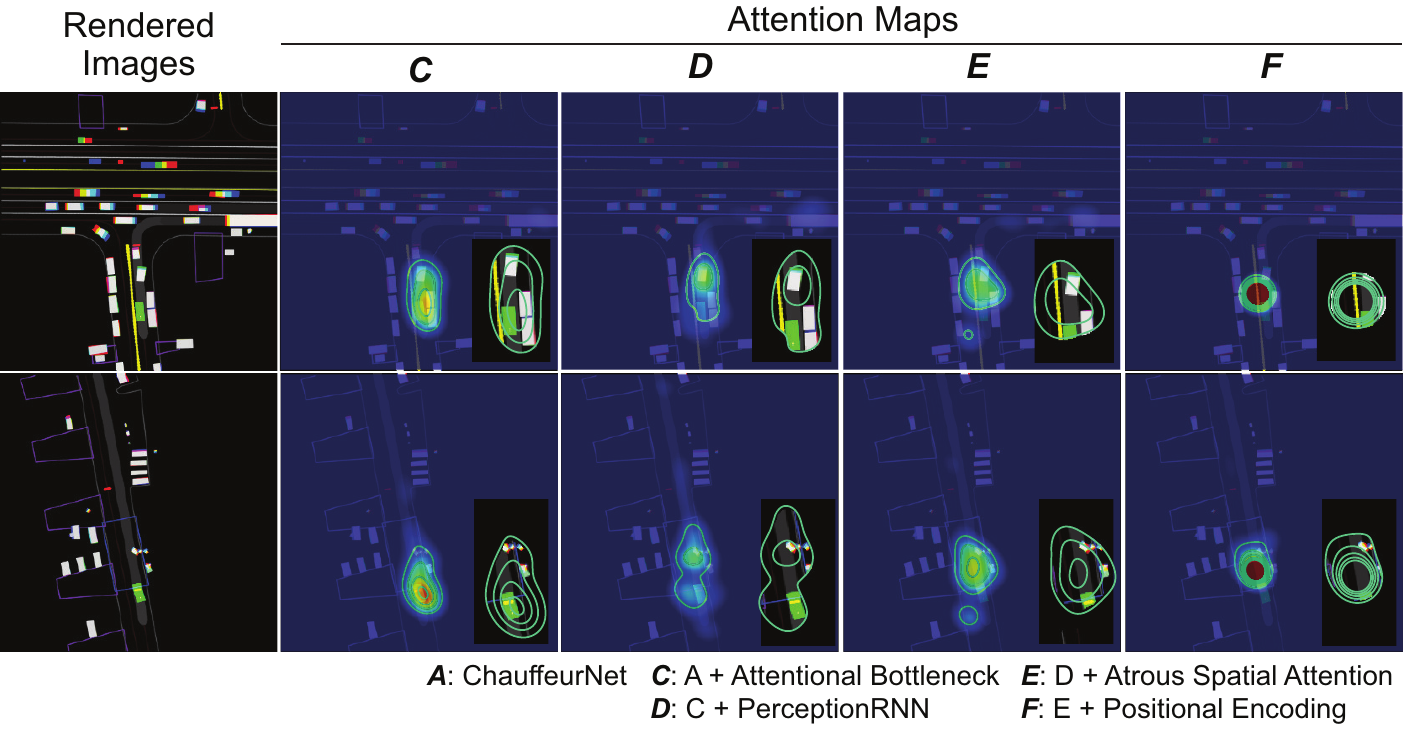}
    \end{center}
    \caption{Qualitative comparison of attention maps between Attentional Bottleneck ablation designs.}
    \label{fig:ablation_imgs} \vspace{-1.3em}
\end{figure}

\subsection{Quantitative Analysis}
We start by quantitatively comparing our attentional bottleneck model (model F) with the baseline ChauffeurNet~\cite{bansal2018chauffeurnet} (model A) and ChauffeurNet with visual attention (model B) models (see Fig.~\ref{fig:diagram}). We observe in Fig.~\ref{fig:ablation} that the incorporation of visual attention for improving the interpretability of the baseline model degrades its performance as measured by the larger ADE and FDE numbers. This is not the case with our attentional bottleneck model where we observe improved ADE and FDE numbers -- possibly due to improved focus by the model on specific causal factors. Examples in Fig.~\ref{fig:diagram} (right) compare our attention maps to those from the visual attention model, and confirm that the latter generates verbose attention heat maps -- finding all potentially salient objects. In contrast, our model provides much sparser attention heat maps which are easier to associate with specific objects or rendered features and are thus easier to interpret. This is evident by comparing their distributions where the attention weights from our model are mostly concentrated around zero probability values (see supplemental figures).

\myparagraph{Effect of incorporating Behavior Prediction.}
As detailed in \cite{bansal2018chauffeurnet}, the prediction of potential future trajectories of dynamic objects in the scene helps the network learn better features for the motion generation task. However, to accomplish this the network would need to attend to all the objects. This renders the attentional bottleneck useless for the objective of exposing only the objects relevant for the primary goal of agent motion generation. In this paper, our focus is on the main motion generator branch, but for completeness, we present one particular architecture choice that allows us to enable {\em PerceptionRNN}~\cite{bansal2018chauffeurnet} while preserving the interpretability of the attentional bottleneck. We add an additional FeatureNet and Atrous Spatial Attention branch that encodes only the dynamic object channels $\Obj$ into attended object features $\A_\Obj$ which are fed into PerceptionRNN. To allow the PerceptionRNN losses to influence the motion generation task, we also inject $\A_\Obj$ into the AgentRNN bottleneck branch by modifying features $\bf{f}_i \in \F_\I$ to $\bf{f}_i^{\prime} = [{\bf{f}}_{i}; {\bf{a}}_{i}]$ where $;$ denotes depth concatenation and $\bf{a}_i \in \A_\Obj$. In Fig.~\ref{fig:ablation}, we compare the metrics from a baseline Attentional Bottleneck model against one that is co-trained with PerceptionRNN in the above setup and observe that this co-training indeed provides improvements in all metrics. As shown in Fig.~\ref{fig:ablation_imgs}, incorporation of PerceptionRNN (model D) improves situation-specific dependence on salient dynamic objects in the scene, i.e. vehicles ahead and pedestrians crossing the crosswalk.

\begin{figure}[!t]
    \centering
    \includegraphics[width=\linewidth]{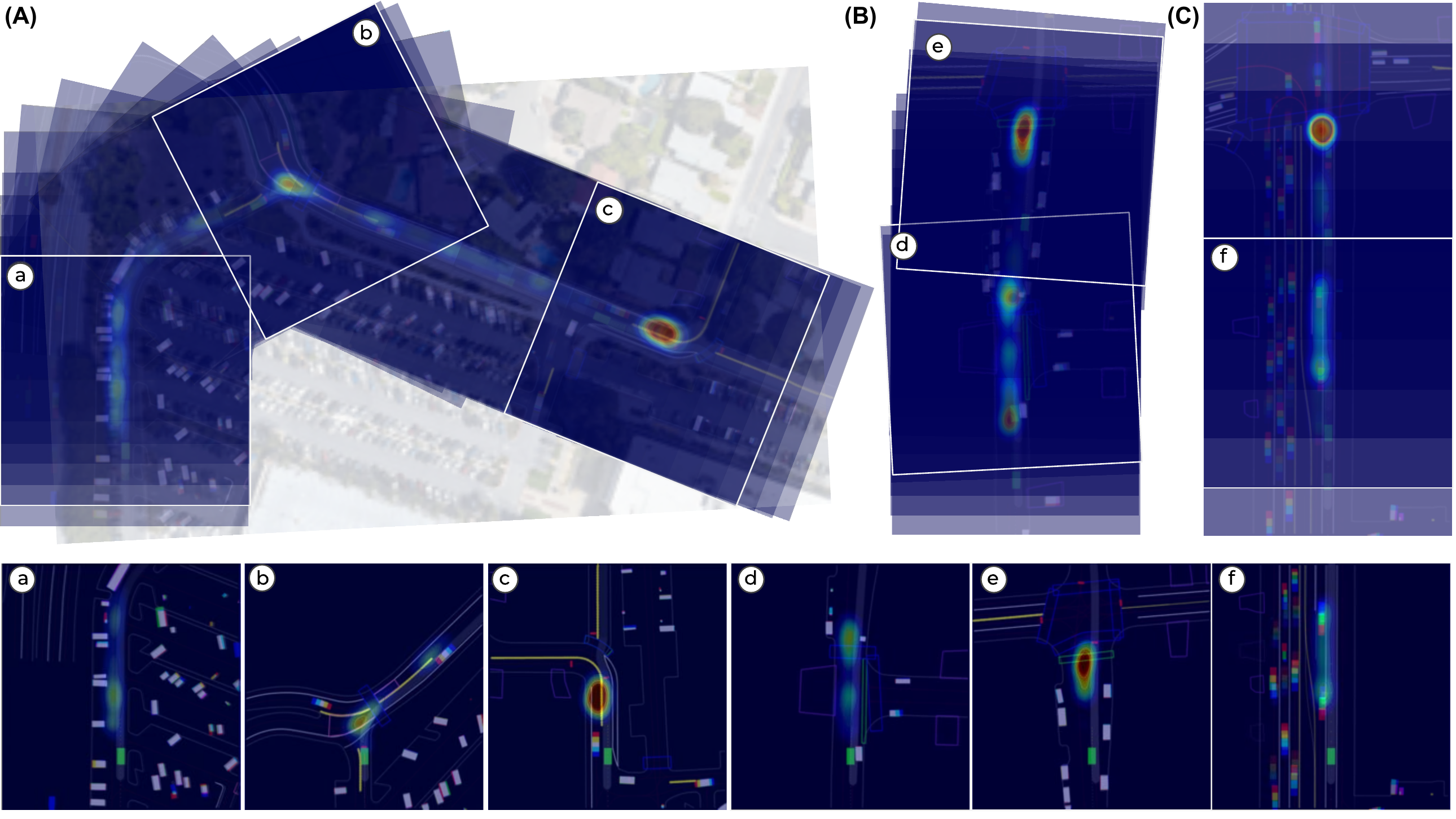}
    \caption{(A-C) Attention maps over time, sampled at every 5 timesteps, illustrate the smooth variation of attention. For better visualization, attention maps are overlaid and shown over the satellite map image. We provide six snapshots on the bottom. Our model appears to attend to important cues, e.g. (a) multiple pinch points, (b) oncoming lanes on a T-junction, (c, e) a stop sign, (d, e) a crosswalk, and (f) vehicles. We also provide a video as supplemental materials. Map data \copyright Google 2020.}
    \label{fig:attn_seq} \vspace{-1.3em}
\end{figure}

\myparagraph{Effect of Atrous Spatial Attention.} 
We illustrate the effect of capturing more non-local information using Atrous convolutions in Fig.~\ref{fig:ablation}. Relative to the model using only a local receptive field, this model achieves better ADE, FDE, and Collision numbers while also improving the sparsity of the attention map. Qualitatively, this model tends to attend to inter-object image regions (compare D vs. E in \cref{fig:ablation_imgs}).

\myparagraph{Effect of Positional Encoding.} 
Quantitatively, the motion generation regression performance slightly degrade (but is still better than the original model) as we concatenate a spatial basis to obtain a latent bottleneck vector. The attention maps become more sparse and we find them to have better spatial alignment with the causal objects (compare E and F in \cref{fig:ablation_imgs}). Thus there is some tension between sparse maps and motion generation performance.

\begin{figure}[!t]
    \centering
    \includegraphics[width=\linewidth]{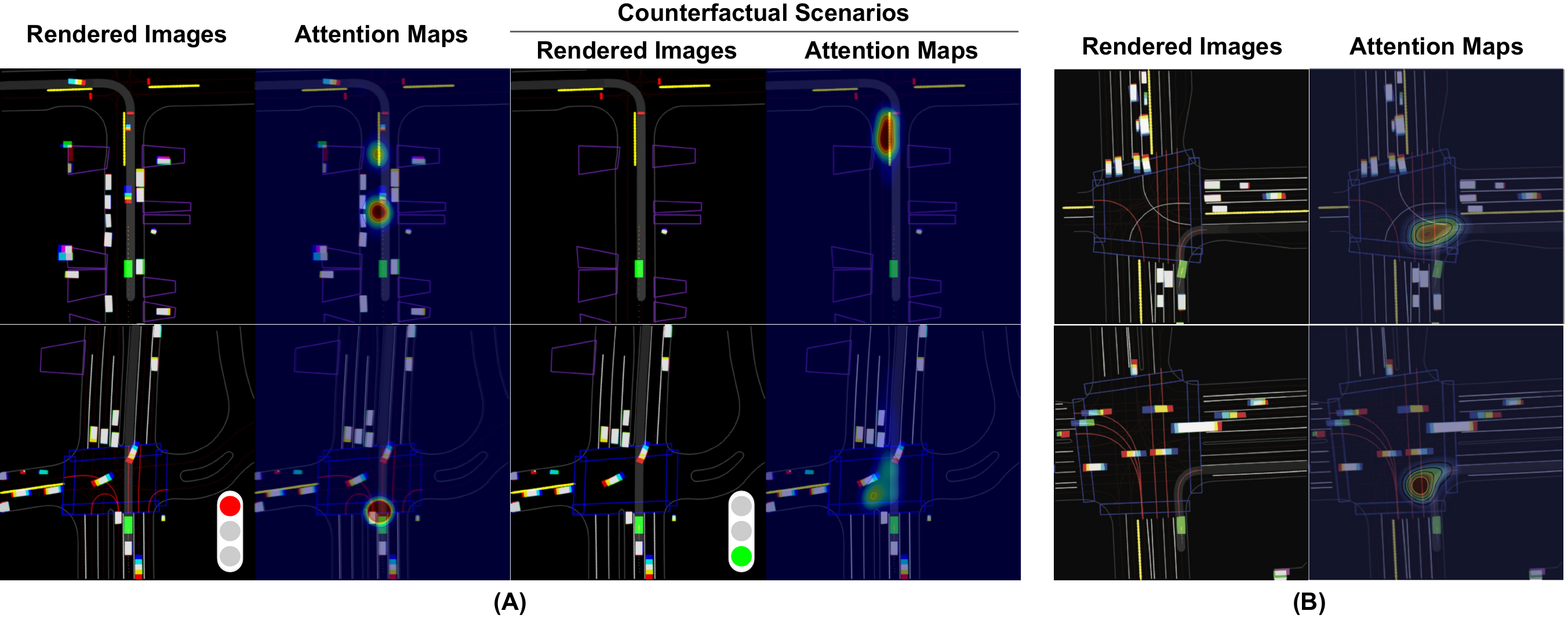}
    \caption{(A) Examples of counter-factual outcomes where the driving model appears to attend alternative important cues, e.g. (top) vehicles ahead $\rightarrow$ a stop sign, (bottom) red lights $\rightarrow$ vehicles ahead. (B) Examples where the driving model appears to under-attend to important cues, e.g. ignoring oncoming lanes on intersections.}
    \label{fig:counterfactual} \vspace{-1.3em}
\end{figure}

\subsection{Qualitative Analysis}
Fig.~\ref{fig:attn} shows several examples covering common driving scenarios that involve slowing down, stopping, avoiding obstacles etc. We visualize a flattened view of the input (odd rows) and the corresponding attention maps (even rows). The attention maps are overlaid on the input images and include contour lines for easier viewing. Note that our attention maps are quite sparse and provide plausible visual evidence of what triggered a particular behavior, e.g. highlighting stop/yield signs or vehicles ahead causing deceleration, road contours when on curved roads, or multiple pinch points from parked vehicles when on narrow roads. 

\myparagraph{Attention Maps over Time.} In Fig.~\ref{fig:attn_seq}, we illustrate attention maps over time for typical driving scenarios, i.e. avoiding multiple pinch points, stopping at an intersection with 4-way stop signs, etc. We also provide snapshots on the right column where our model attends to important driving-related cues. Our supplemental video also demonstrates the smooth variation of the attention maps across several temporal sequences illustrating the changing causality with each driving decision.

\myparagraph{Counterfactual Experiments.}
Fig.~\ref{fig:counterfactual} (A) shows examples where the attention maps change in the counterfactual driving scenarios by getting rid of a subset of environmental descriptions, i.e. dynamic objects (i.e. with vs. without vehicles) and traffic light states (i.e. red vs. green). These indicate that our model successfully focuses on the correct cues in specific situations. We provide more diverse examples as supplemental materials.

\myparagraph{Failure Cases.}
Fig.~\ref{fig:counterfactual} (B) shows examples where the attention maps fail to capture the correct causal behavior. This indicates either that the original model failed to focus on the correct cues in specific situations (e.g. ``look both ways before making a right turn''), or that the attention model is not able to correctly explain this situation.

\section{Conclusions}
We described an approach for improving interpretablity of a mid-to-mid deep driving model by augmenting a visual attention model with an attentional bottleneck layer. Our results highlight sparse attention maps which are easy to interpret and do not degrade model performance. We see opportunity in taking this further to generate instance level attention maps and to also use these maps as a guide to improving the performance of the baseline driving model.

\section*{Acknowledgements}
We thank Dragomir Anguelov, Anca Dragan, and Alexander Gorban at Waymo Research, John Canny, Trevor Darrell, Anna Rohrbach, and Yang Gao at UC Berkeley for their helpful comments.

\bibliographystyle{splncs04}
\bibliography{manuscript}
\end{document}